\pdfoutput=1

\documentclass[11pt]{article}

\usepackage[]{acl}

\usepackage{times}
\usepackage{latexsym}
\usepackage[T1]{fontenc}
\usepackage[utf8]{inputenc}
\usepackage{microtype}
\usepackage{multicol}
\usepackage{multirow}
\usepackage{booktabs}
\usepackage{graphicx}
\usepackage{booktabs}
\usepackage{multirow}
\usepackage{tabularx}
\usepackage{hyperref}

\newcommand{\tb}[1]{{\textbf{#1}}}
\usepackage{makecell}
\usepackage{colortbl}
\usepackage{amssymb}
\usepackage{caption}
\usepackage{subcaption}

\definecolor{lightred}{HTML}{f4dee0} 
\definecolor{lightgreen}{HTML}{ecf4eb} 
\newcommand{\increasebg}[1]{{\cellcolor[HTML]{f4dee0}{#1}}}
\newcommand{\smallincreasebg}[1]{{\cellcolor[HTML]{faeeef}{#1}}}
\newcommand{\decreasebg}[1]{{\cellcolor[HTML]{ecf4eb}{#1}}}

\newcommand{\para}[1]{\vspace{.05in}\noindent\textbf{#1}}
\newcommand{\cellcenter}[1]{\multirow{1}{*}[-0.5ex]{#1}}
\usepackage[T1]{fontenc}
\usepackage[utf8]{inputenc}
\usepackage{microtype}
\usepackage{inconsolata}

\title{Bridging Cultural Nuances in Dialogue Agents through \\Cultural Value Surveys}

\newcommand{\hust}{$^{1,}$}
\newcommand{\ku}{$^2$}
\newcommand{\scut}{$^3$}

\author{Yong Cao\hust\ku, Min Chen\scut, Daniel Hershcovich\ku \\
{\hust}Huazhong University of Science and Technology \\
{\ku}Department of Computer Science, University of Copenhagen \\
{\scut}School of Computer Science and Engineering, South China University of Technology \\
\normalsize{\texttt{yongcao\_epic@hust.edu.cn, minchen@ieee.org, dh@di.ku.dk}}}

\begin{document}
\maketitle
\begin{abstract}
The cultural landscape of interactions with dialogue agents is a compelling yet relatively unexplored territory. It's clear that various sociocultural aspects---from communication styles and beliefs to shared metaphors and knowledge---profoundly impact these interactions. To delve deeper into this dynamic, we introduce cuDialog, a first-of-its-kind benchmark for dialogue generation with a cultural lens. We also develop baseline models capable of extracting cultural attributes from dialogue exchanges, with the goal of enhancing the predictive accuracy and quality of dialogue agents. To effectively co-learn cultural understanding and multi-turn dialogue predictions, we propose to incorporate cultural dimensions with dialogue encoding features. Our experimental findings highlight that incorporating cultural value surveys boosts alignment with references and cultural markers, demonstrating its considerable influence on personalization and dialogue quality. To facilitate further exploration in this exciting domain, we publish our benchmark publicly accessible at \url{https://github.com/yongcaoplus/cuDialog}.
\end{abstract}

\section{Introduction}
Culture can be defined as the combinations of beliefs, norms, and customs among groups \cite{tomlinson-etal-2014-capturing}. Implicit cultural cues hinted in dialogue utterances reveal different values and beliefs among speakers, which reflects their way of thinking \cite{nisbett2001culture} and emotions \cite{almuhailib2019analyzing, sun-etal-2021-cross, ma-etal-2022-encbp}. 
While pre-trained language models (PLMs) have shown impressive performance on dialogue tasks \cite{gu-etal-2021-detecting, liu-etal-2021-visually, sweed-shahaf-2021-catchphrase}, 
their cultural bias in terms of values and their inconsistency in many other cultural aspects \cite{fraser-etal-2022-moral} has severe implications on the prospect of employing them for interaction with speakers of diverse cultural backgrounds \cite{hershcovich-etal-2022-challenges}. 
This is particularly crucial in the context of culturally-related topics~\cite{zhou-etal-2023-cross, zhou-etal-2023-cultural}, where acknowledging and understanding cultural differences becomes essential.
For example, scholars tend to believe that Eastern societies have a more communal or collective orientation compared to that Western societies \cite{lomas2023complexifying}.

\begin{figure}[t]
	\centering
	\includegraphics[width=1.0\columnwidth]{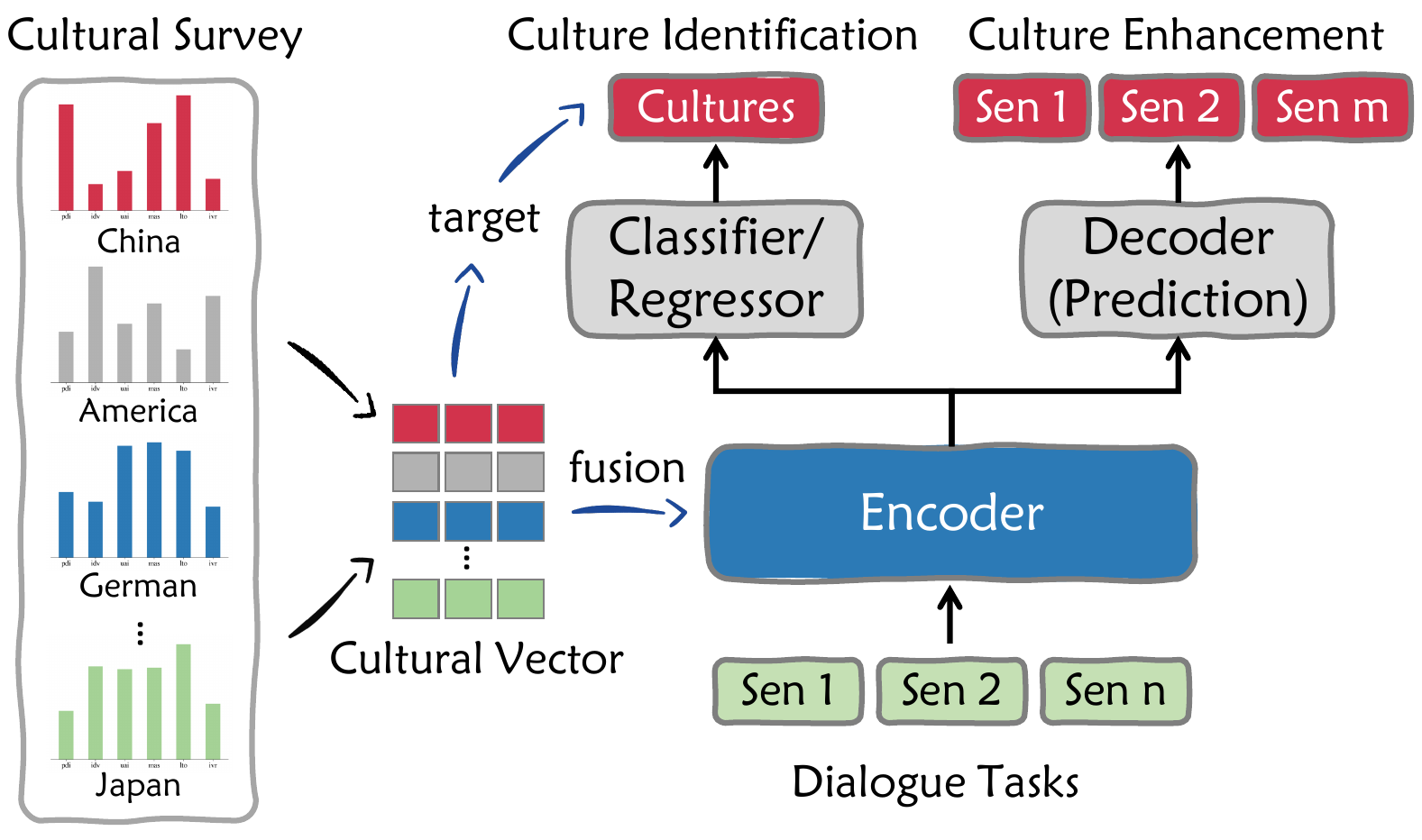}
	\caption{Our proposed framework: Utilizing cultural survey \cite{hofstede1984culture} as an additional vector for multi-turn dialogue culture identification and dialogue prediction enhancement, leveraging our proposed multicultural dialogue benchmark dataset, \textit{cuDialog}.}
	\label{fig:figure1}
\end{figure}

Previous studies in the field of cross-cultural NLP \cite{arora-etal-2023-probing, hammerl2022speaking, johnson2022ghost, santurkar2023opinions} have primarily utilized probing methods to study the characteristics of models or agents. For instance, \citet{cao-etal-2023-assessing} applied the Hofstede Culture Survey \cite[][see \S\ref{sec:dim}]{hofstede1984culture} to probe ChatGPT, a prominent dialogue system, revealing a distinct disparity between the system and human society. This underscores the need to enhance dialogue agents' performance by incorporating cultural dimensions. However, developing culturally adaptive dialogue agents poses a significant challenge due to the scarcity of suitable datasets. While there are available multicultural corpora focused on specific domain tasks such as news \cite{ma-etal-2022-encbp} and image captions \cite{liu-etal-2021-visually}, there is currently a lack of datasets specifically designed for cross-cultural dialogue tasks.

To address this research gap, we introduce cuDialog, an extensive English-language benchmark for multicultural dialogues. Our benchmark covers 13 cultures and 5 genres, specifically designed to mitigate the impact of linguistic variations and emphasize implicit cultural cues. Within cuDialog, we propose two culture understanding tasks and one dialogue generation task, offering a comprehensive framework for evaluating and advancing cultural understanding in dialogue systems.

Specifically, as depicted in Figure \ref{fig:figure1}, we design several baselines on culture classification and regression tasks, showing that cultural attributes behind dialogues can be identified. We leverage the soft cultural knowledge provided by the Hofstede Culture Survey \cite{hofstede1984culture}, which defines six cultural dimensions to measure the cultural attributes of different countries and provides statistical results for numerous nations. To utilize this external knowledge, we present a novel feature fusion mechanism based on an encoder-decoder generation framework, by considering using culture to assist separability in dialog generation. Experimental results reveal that incorporating cultural value representation can improve alignment with references, indicating better cultural representation.

In summary, our contributions are as follows:
(1) We introduce cuDialog, a multicultural dialogue benchmark dataset specifically tailored to different genres, enriched with cultural survey annotations.
(2) We develop several baseline models that effectively capture cultural nuances and propose three dialogue tasks.
(3) We demonstrate the feasibility of capturing cultural nuances and the impact of incorporating cultural representation into dialogue systems, highlighting the significance of considering cultural differences in dialogue modeling.

\section{Related Work}
\paragraph{Culture-oriented benchmarks.}
Researchers have developed a range of culture-oriented benchmarks to investigate the impact of culture on language understanding and generation tasks. These benchmarks involve collecting and annotating multilingual and multicultural corpora to study cultural effects in downstream tasks. For instance, benchmarks have been introduced for news classification across different countries \cite{ma-etal-2022-encbp} and for analyzing user statements reflecting different cultures using text and images \cite{liu-etal-2021-visually}. Other benchmarks focus on detecting culture differences and user attributes, spanning both small-scale \cite{sweed-shahaf-2021-catchphrase} and large-scale \cite{qian2021pchatbot} datasets. Furthermore, recent works have explored in-domain cross-cultural benchmarks, such as multilingual moral understanding and generation \cite{guan-etal-2022-corpus}, and culture-specific time expression grounding \cite{shwartz-2022-good}. While \citet{zhang2022mdia} proposed a multilingual conversation dataset, it lacks cultural annotations.

\paragraph{Cultural attributes learning.}
Traditional approaches for capturing cultural differences often rely on probabilistic models, such as Latent Dirichlet Allocation \cite{pennacchiotti2011democrats, al2012homophily, tomlinson-etal-2014-capturing}. However, the emergence of unsupervised learning and advancements in pre-trained language models (PLMs) have sparked interest in utilizing PLMs to learn cultural attributes and user profiles \cite{gu-etal-2021-detecting, fraser-etal-2022-moral}.

\paragraph{Culture-sensitive dialogue agents.}
Previous studies \cite{tomlinson-etal-2014-capturing, ma-etal-2022-encbp} have demonstrated the benefits of equipping dialogue agents with an understanding of cultural differences for natural language understanding (NLU) and generation (NLG) tasks, even in general natural language processing tasks. For example, \citet{fu-etal-2022-thousand} proposed the use of a persona-specific memory network to jointly encode cultural background and user profiles, enhancing the NLG task for dialogue agents. \citet{kanclerz-etal-2021-controversy} introduced personalized approaches that respect individual beliefs expressed through user annotations. Additionally, \citet{wu-etal-2021-personalized} incorporated user queries, cultural-related comments, and user profiles as encoded features to generate personalized responses, demonstrating the efficacy of leveraging both features in improving dialogue agent satisfaction. Moreover, leveraging external knowledge by retrieving user-related cultural and attribute documents has shown promising improvements, providing additional guidance for model training \cite{majumder-etal-2021-unsupervised, guan-etal-2022-corpus}. These works collectively highlight the value of incorporating cultural aspects into dialogue systems and leveraging personalized approaches for more effective and satisfactory interactions.  While recent efforts have incorporated commonsense knowledge \cite{varshney-etal-2022-commonsense} and socio-cultural norms \cite{moghimifar2023normmark} into dialogue agents, these approaches have primarily focused on monocultural settings, neglecting the broader context of multicultural dialogue.


\begin{figure}[t]
	\centering
	\includegraphics[width=1.0\columnwidth]{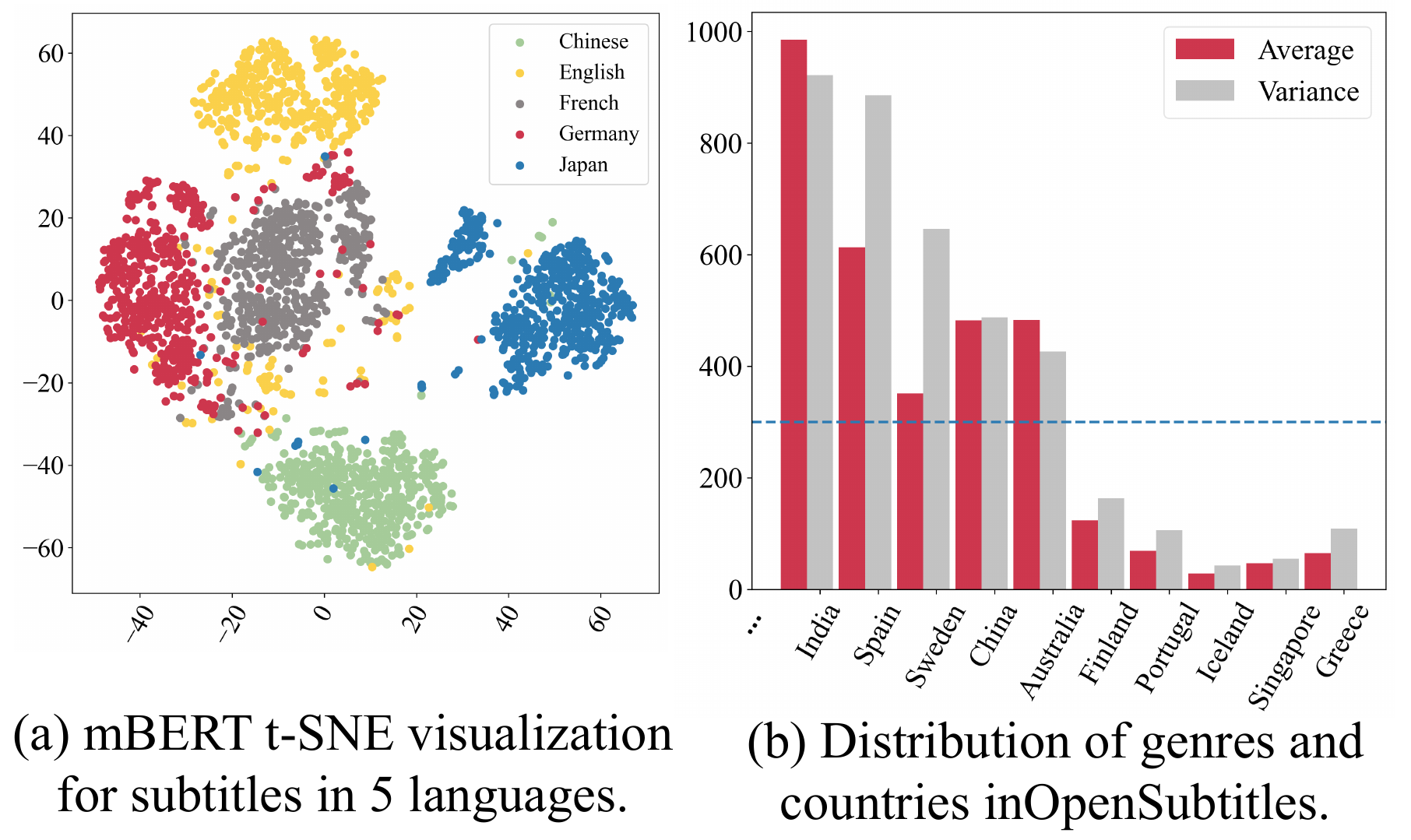}
	\caption{Corpus distribution.}
	\label{fig:genres}
\end{figure}

\section{Cultural Dimensions}\label{sec:dim}
The Hofstede Culture Survey \cite{hofstede1984culture} identifies six cultural dimensions that capture different aspects of cultural values:

\textbf{Power Distance (pdi):} Reflects the acceptance of unequal power distribution within a society.

\textbf{Individualism (idv):} Measures the level of interdependence versus self-definition within a culture.

\textbf{Masculinity (mas):} Examines the emphasis on competition, achievement, and assertiveness versus caring for others and quality of life.

\textbf{Uncertainty Avoidance (uai):} Deals with response to ambiguity and minimizing uncertainty.

\textbf{Long-Term Orientation (lto):} Describes how cultures balance tradition with future readiness.

\textbf{Indulgence (ivr):} Focuses on the control of desires and impulses based on cultural upbringing.

These dimensions offer valuable insights into the beliefs, behaviors, and attitudes that vary across societies. By incorporating these dimensions in our dataset for the corresponding countries, we provide a benchmark for evaluating the ability of dialogue systems to capture and adapt to cultural nuances. This enables researchers to assess the cultural sensitivity and adaptability of dialogue systems in a standardized manner. The survey results, freely available online for 111 countries,\footnote{\url{https://geerthofstede.com/research-and-vsm/dimension-data-matrix/}} serve as a valuable resource for integrating cultural dimensions into dialogue system enhancement and evaluation.

\section{Multicultural Dialogue Dataset}
In this section, we introduce the collection, benchmarking, and statistics of our proposed multicultural dataset. The cuDialog dataset contains four components: histories, golden predictions, culture label, and culture dimension scores, serving our proposed tasks, including culture classification, cultural alignment and dialogue generation, etc. 

\paragraph{Data source.} 
We gather multicultural dialogues from the OpenSubtitles 2018 dataset\footnote{\url{https://opus.nlpl.eu/OpenSubtitles-v2018.php}} \cite{lison-etal-2018-opensubtitles2018}, which comprises a vast collection of subtitles extracted from movies and television shows. The OpenSubtitles 2018 dataset offers extensive coverage of multiple languages, providing subtitle data in text format that is well-suited for training and evaluating a diverse range of NLP models. With its inclusion of various genres, such as action, drama, comedy, and documentaries, the dataset ensures an inclusive representation of linguistic styles and domains. While the dataset has been widely utilized in language identification \cite{toftrup-etal-2021-reproduction}, domain adaptation \cite{thompson-etal-2019-overcoming, lai-etal-2022-4}, and machine translation \cite{costa2022no, zhang-ao-2022-yitrans}, it is essential to recognize that it also contains substantial cultural cues. To our knowledge, our work represents the first application of this dataset for culture-focused research, complemented by cultural annotations.

\begin{figure}[t]
	\centering
	\includegraphics[width=0.85\columnwidth]{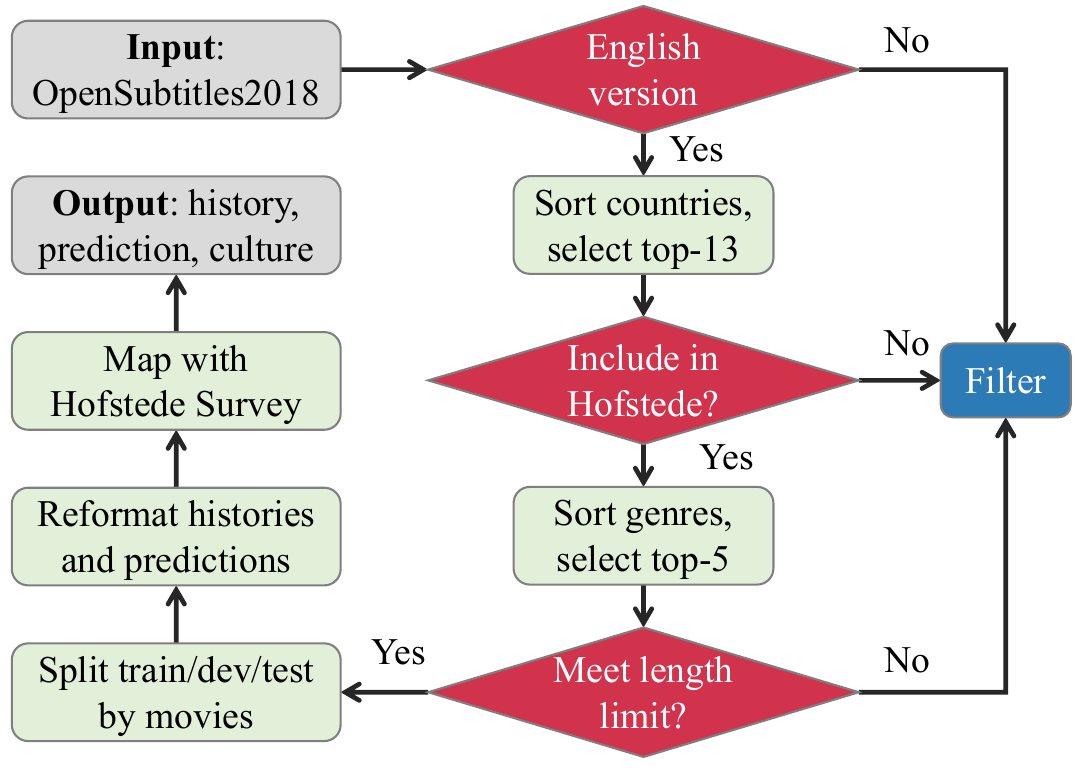}
	\caption{The pipeline of the cuDialog dataset construction process with our designed filtering strategy.}
	\label{fig:data_construct}
\end{figure}

\paragraph{Language and culture selection.} 
Our research aims to explore the cultural differences underlying linguistic variations. We acknowledge that linguistic variations themselves serve as strong cultural features, which can have an impact on aspects such as common grounding and beliefs. To investigate the cultural cues related to beliefs and values, we conducted an analysis using a subset of 500 randomly extracted samples from the OpenSubtitles dataset. These samples were encoded by mBERT and visualized using the t-SNE method \cite{van2008visualizing}, with a specific focus on the representation of data from five distinct countries. The visualization revealed distinct separations in the representation space based on different languages, making it challenging to capture cultural cues beyond linguistic variations. This motivated our decision to utilize English subtitles, as they exhibit less trivial separability (Figure \ref{fig:genres}a). As a result, our benchmark dataset universally employs English subtitles that encompass all cultures. The English subtitles in our dataset comprise both human-translated and machine-translated versions.

Furthermore, to establish a comprehensive benchmark dataset, we analyzed various genres and countries (as depicted in Figure \ref{fig:genres}b). We selected the top-five genres, namely action, comedy, drama, romance, and crime, as the basis for our dataset. In terms of country selection, we established a threshold of at least 50 movies per genre, ranked all countries accordingly, and chose the top-13 countries to represent cultures in our dataset. These countries include the USA, UK, France, Japan, Germany, Canada, Italy, South Korea, India, Spain, Australia, China, and Sweden.

\begin{table}
\centering
\resizebox{0.5\textwidth}{!}{
\begin{tabular}{c|ccccc}
\toprule
\multirow{2}{*}{\textbf{Set}} & \multicolumn{5}{c}{\textbf{Genres}} \\ \cline{2-6} 
                         & \cellcenter{\textbf{Action}} & \cellcenter{\textbf{Comedy}} & \cellcenter{\textbf{Drama}} & \cellcenter{\textbf{Romance}} & \cellcenter{\textbf{Crime}} \\ \midrule\midrule
                         
\multicolumn{6}{c}{\textit{\textbf{Samples}}} \\ \midrule
\multicolumn{1}{l|}{Train}       & 108,934 & 137,475 & 109,534 & 114,467 & 112,695  \\
\multicolumn{1}{l|}{Dev}          & 12,808 & 17,361 & 13,256 & 13,697 & 14,313  \\
\multicolumn{1}{l|}{Test}         & 15,336 & 18,213 & 16,179 & 15,705 & 16,853  \\ \midrule
\multicolumn{6}{c}{\textit{\textbf{Movies}}}      \\ \midrule
\multicolumn{1}{l|}{Train}       & 728 & 728 & 728 & 728 & 728  \\   
\multicolumn{1}{l|}{Dev}          & 91 & 91 & 91 & 91 & 91  \\
\multicolumn{1}{l|}{Test}         & 104 & 104 & 104 & 104 & 104  \\ \midrule
\multicolumn{6}{c}{\textit{\textbf{Tokens}}} \\ \midrule
\multicolumn{1}{l|}{Vocab}  & 34,883 & 36,292 & 32,566 & 32,666 & 33,416  \\
\multicolumn{1}{l|}{\#Avg}  & 71.15 & 70.32 & 72.38 & 71.42 & 72.05  \\
\bottomrule
\end{tabular}}
\caption{The statistics of cuDialog. Here we split train, dev and test set by movies to avoid data leakage. \#Avg is the average number of tokens by mT5 tokenizer. Vocab is the total vocabulary size.}
\label{tb:statistics_cuDialog}
\end{table}

\paragraph{Pipeline.}
Our cuDialog dataset construction pipeline (Figure \ref{fig:data_construct}) involves gathering a comprehensive movie category index and extracting the corpus from each movie. We create multi-turn dialogues to capture cultural cues, with each dialogue containing eight sentences. These dialogues are divided into an input history $Q_i$ (first five sentences) and prediction references $R_i$ (last three sentences). Each dialogue is labeled with a cultural label $C_i$ representing the country of origin, and cultural value scores $S_i$ (\S\ref{sec:dim}) are assigned accordingly.

\paragraph{Dialogue format.}
The cuDialog dataset is represented as $\{d_i \in D | d_i = (Q_i, R_i, C_i, S_i)\}$. An example of a dialogue in the cuDialog format is presented in Table \ref{tb:case_cuDialog}. To ensure data quality, we remove short contexts and responses that provide limited information, making it challenging for dialogue agents to infer the cultural background effectively. Additionally, we eliminate emojis and address encoding errors to enhance overall quality.

\begin{table}
\centering
\begin{tabularx}{\columnwidth}{X}
\toprule
\small \textbf{History:}
\textit{His mortal flesh belonged to the fire, his immortal soul to the flames of Hell. $\big\vert$ A gag blocked his mouth. $\big\vert$ You'd have thought it was a corpse being led to its grave, $\big\vert$ ``yet it was a living man whose torments were to gruesomely entertain the people.'' $\big\vert$ Forgive me, I'll break off here.} \\
\small
\textbf{Golden Predictions:}
\textit{Will you amuse us now with details of an execution during the Inquisition? $\big\vert$ No, I beg your pardon. $\big\vert$ I'm deeply impressed.} \\
\midrule
\small
\textbf{Culture:} Germany. \\
\small
\textbf{Culture Score: } 35, 67, 66, 65, 83, 40. \\
\bottomrule
\end{tabularx}
\caption{A \textit{Romance} genre example from cuDialog with four fields: multi-turn history, golden predictions, culture category, and cultural value dimension scores.}
\label{tb:case_cuDialog}
\end{table}

\begin{figure*}[t]
	\centering
	\includegraphics[width=1.0\textwidth]{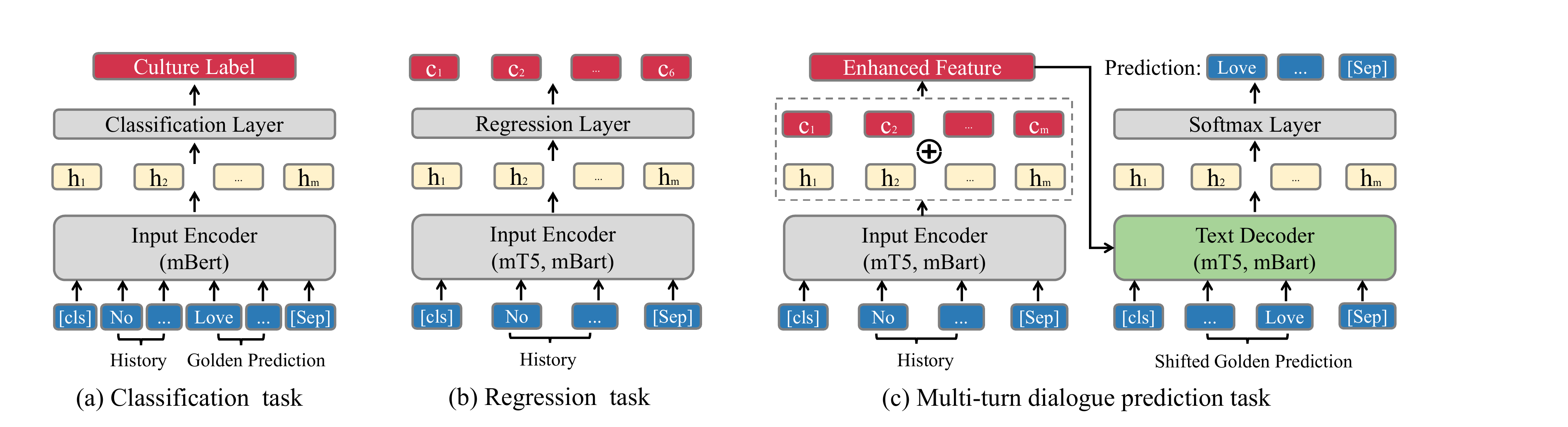}
	\caption{Cultural features enhancement in dialogue tasks using the Encoder-Decoder framework with our proposed benchmark dataset, i.e., cuDialog. Our novelty lies in the cultural aspects which we highlight in red, employing culture vectors as training targets and additional features. $\oplus$ denotes padding and fusion strategy.}
	\label{fig:network}
\end{figure*}

\paragraph{Dataset statistics.}
To facilitate comparative analysis and maintain dataset balance, we ensure a consistent number of movies across different genres. Table \ref{tb:statistics_cuDialog} presents an overview of the cuDialog dataset's statistics. Each genre comprises approximately 130 to 160 thousand dialogues, with a total of 923 movies and an average sentence length of around 71, considering both the input histories and prediction references. The dataset is divided into train (80\%), validation (10\%), and test (10\%) sets, with no overlap between movies in the test set and those in the train set. This partitioning is performed at the movie and television show level, enabling dialogue-related tasks.\footnote{More detailed dataset statistics are in Appendix \ref{ax:hosfeted_survey}.}

\section{Cultural Dialogue Tasks}
Drawing from the insights gained from previous research \cite{arora-etal-2023-probing, cao-etal-2023-assessing}, which highlighted the challenges faced by pre-trained models and dialogue agents in capturing cultural differences, we aim to analyze cultural attributes and explore effective mechanisms for cultural alignment. We pose the following research questions:

\begin{itemize}
    \item \textbf{RQ1:} Can our cuDialog dataset effectively capture and identify cultural dimensions?
    \item \textbf{RQ2:} How do cultural nuances impact the performance of dialogue agents across cultures?
\end{itemize}

To address these research questions, we introduce three dialogue tasks, depicted in Figure \ref{fig:network}.

To address RQ1, we go beyond the conventional approach and examine whether the dialogues in cuDialog exhibit discernible cultural differences that can be effectively classified. Our first task, \textbf{culture classification}, delves into the identification of cultural variations in the dataset. Additionally, we explore the \textbf{cultural dimension score regression} task to investigate the feasibility of inferring fine-grained cultural labels. These tasks necessitate capturing cross-cultural differences and exploit the multicultural variety of cuDialog.

To tackle RQ2, we propose a \textbf{multi-turn dialogue prediction} task based on the hypothesis of cultural separability. By incorporating cultural features into the dialogue agent framework, we aim to enhance the performance of dialogue agents by considering the influence of cultural nuances. This task provides valuable insights into how culture impacts dialogue systems and sheds light on the role of cultural factors in improving the overall performance and adaptability of dialogue agents.

\subsection{Culture Classification}
In the culture classification task, depicted in Figure \ref{fig:network}(a), the goal is to predict the correct culture label $C_i$ among the 13 countries, given a dialogue history $Q_i$ and golden prediction $R_i$. The task involves predicting $\mathcal{P}_c(c|h_i,r_i)$, where $c \in C_i$, $h_i \in Q_i$, and $r_i \in R_i$. Notably, the input contains the query and response as a combined context. We specifically choose the multi-turn dialogue format instead of single-turn dialogues due to the short and limited information present in OpenSubtitles sentences. By ensuring longer text, we aim to capture and learn the cultural cues effectively. This task can be modeled using encoder-only models and does not involve generation or address cultural dimensions.

\subsection{Cultural Dimension Regression}\label{sec:regression}
In cultural dimension regression, we leverage the cultural dimensions obtained from the Hofstede Culture Survey (\S\ref{sec:dim}) as fine-grained cultural labels. As depicted in Figure \ref{fig:network}(b), we employ a regression layer that operates on the encoder hidden states to predict the six-dimensional cultural scores for each dialogue. Specifically, we aim to predict $\hat{\mathcal{P}}_c(\hat{c}|h_i)$, where $\hat{c}$ represents the six-dimensional cultural vectors and $\hat{\mathcal{P}}_c$ denotes the prediction. In this task we use only the history text instead of concatenating the history and golden predictions. This adjustment allows us to effectively capture the cultural dimensions and assess their impact on dialogue systems' performance, using encoder-decoder models.

\subsection{Multi-Turn Dialogue Prediction}
Culture plays a crucial role in dialogue generation, as it influences the choice of words, expressions, and behaviors in conversations. To capture the cultural nuances and ensure culturally appropriate responses, we propose a multi-turn dialogue prediction task that incorporates cultural value representations.
In our approach, we utilize the cultural dimensions (\S\ref{sec:dim}) as representations of cultural values. These dimensions serve as contextual cues that guide the dialogue generation process by integrating them into the encoder-decoder framework.

In this task, we employ an encoder-decoder framework, where the encoder processes the dialogue history $h_i$ to obtain the hidden states $\mathcal{H}^{(1)}, ..., \mathcal{H}^{(L)}$. We consider the cultural dimensions $\hat{c}$ obtained from a culture regression model (\S\ref{sec:regression}) as representations of cultural values. To incorporate these dimensions into the dialogue generation process, we extend each dimension to match the length of the hidden states, resulting in $\hat{c}_d$. We concatenate $\hat{c}_d$ with the hidden states at each layer:
\begin{equation}
    \mathcal{H}^{(1)}_d, ..., \mathcal{H}^{(L)}_d = \mathcal{D} (\mathcal{H}^{(1)}, ..., \mathcal{H}^{(L)}, \hat{c}_d)
\end{equation}

Finally, the decoder generates the predicted response by utilizing the concatenated hidden states. 

This approach requires the model to consider cultural dimensions, ensuring that the generated responses align with the underlying cultural values.

\section{Experiments}

\subsection{Evaluated Models}
To extensively evaluate the performance of currently available models, we select various models for evaluation, encompassing both encoder and encoder-decoder frameworks, as well as monolingual and multilingual models. Specifically, we evaluate the following baselines for culture classification tasks: BERT \cite{devlin-etal-2019-bert}, multilingual BERT, RoBERTa \cite{liu2019roberta}, and XLM-RoBERTa \cite{conneau-etal-2020-unsupervised}. For the culture regression task, we evaluate T5 \cite{raffel2020exploring}, mT5 \cite{xue-etal-2021-mt5}, BART \cite{lewis-etal-2020-bart}, and mBART50 \cite{tang2020multilingual}. For dialogue prediction, we evaluate mT5 on five genres.

\subsection{Experimental Setup}
Using pre-trained models from HuggingFace \cite{wolf-etal-2020-transformers},\footnote{See Appendix \ref{ax:pre_trained_models} for full model identifiers.} we use one A100 GPU for culture classification and regression and two A100 GPUs for multi-turn dialogue prediction. As hyperparameters, we set the batch size to 128, 256, and 64 for culture classification, regression, and prediction tasks, respectively. We use an early stopping strategy with a patience of 2 or 3. For generation, we employ beam search with a width of 3, temperature of 0.7, and repetition penalty of 1.2.\footnote{More details for reproducibility are in Appendix \ref{ax:hyper_parameters}.}

\subsection{Evaluation Metrics}
The evaluation metrics used in our study depend on the task at hand. For classification tasks, we employ recall, precision, and F1 score. Regression tasks are evaluated using the Spearman correlation coefficient, R2 score, and root mean squared error (RMSE). For generation, we use BLEU measuring n-gram overlap, ROUGE-L considering the longest common subsequence, BERTScore assessing similarity using contextualized embeddings, and Distinction evaluating distinctiveness in terms of diversity and uniqueness. These metrics align with the approach proposed by \citet{zhang2022mdia}.

\subsection{Main Results}

\paragraph{Culture Classification.} 
Table \ref{tb:classificaton_results} presents the results for culture classification, comparing the performance of monolingual models (BERT and RoBERTa) with multilingual models (mBERT and XLM-R).\footnote{Additional scores for each culture are in Appendix \ref{ax:classification_ret}.} Interestingly, we observe that the monolingual models demonstrate superior performance in this task, suggesting a slight disadvantage for multilingual models within the context of an English corpus encompassing all cultures. It is noteworthy that the action and crime genres exhibit a higher suitability for culture classification, aligning with our expectations. This can be attributed to the significant cultural variations in the interpretation of criminal activities, such as the legality of firearm possession \cite{boine2020gun}.

In contrast, the comedy corpus performs relatively poorly in culture classification, which can be attributed to the challenges of translation. Prior research \cite{jiang2019cultural} has indicated the existence of cultural differences in humor usage between Eastern and Western societies. Western cultures tend to associate humor with positivity and view it as a natural form of amusement expression \cite{martin2018psychology}, whereas Eastern cultures often hold contrasting attitudes towards humor \cite{dong2010exploration}. However, we contend that during the translation process, a significant number of comedic elements lose their impact, resulting in diminished distinction for the models.

\begin{table}[]
\centering
\resizebox{0.48\textwidth}{!}{
\begin{tabular}{l|ccccc}
\toprule 
Model & Action & Comedy & Drama & Romance & Crime \\
\midrule\midrule
RoBERTa & 87.93 & 75.43 & 82.39 & 83.20 & 85.29 \\
XLM-R & 86.50 & 75.39 & 80.69 & 79.29 & 84.27 \\  \midrule
BERT & \tb{\increasebg{88.49}} & \tb{\increasebg{76.80}} & \tb{\increasebg{83.77}} & \tb{\increasebg{82.60}} & \tb{\increasebg{85.70}} \\ 
mBERT & 86.05 & 76.26 & 82.62 & 80.48 & 81.21 \\  
\bottomrule
\end{tabular}}
\caption{F1 scores of dialogue culture classification models for 13 cultural categories. The English-only models RoBERTa and BERT outperform the multilingual models mBERT and XLM-R.}
\label{tb:classificaton_results}
\end{table}

\paragraph{Culture Regression.}
Table \ref{tb:reg_ret} presents the results for culture regression using T5, mT5 , BART and mBART models. We fine-tune the models individually for each genre and compare the alignment between our predictions and human surveys using all 13 culture vectors. We first fine-tune the monolingual models T5 and BART, observing these models demonstrate limited culture alignment capabilities, resulting in poor performance across all evaluation metrics. In contrast, after fine-tuning multilingual models, we observe a significant improvement in cultural alignment. Particularly, mBART outperforms all other models on all tasks, indicating its ability to align with cultural values. This difference in performance can be attributed to the distinct pre-training corpora and tasks employed by each model, and highlight the importance of pre-training tasks in shaping the models' performance and their capacity for cultural alignment.

\begin{table}[t]
\centering
\renewcommand{\arraystretch}{.8}
\resizebox{0.48\textwidth}{!}{
\begin{tabular}{l|ccccc}
\toprule
Method & Action & Comedy & Drama & Romance & Crime \\ \midrule \midrule
\multicolumn{6}{c}{\textbf{\textit{Spearman correlations (COR) $\uparrow$}}} \\
T5 & -0.0321* & 0.0784* & -0.0436* & -0.1144* & -0.0989* \\
mT5 &  \tb\smallincreasebg{0.8135*}   &  \tb\smallincreasebg{0.7432*}  &   \tb\smallincreasebg{0.7825*}    &   \tb\smallincreasebg{0.6919*}           &   \tb\smallincreasebg{0.7757*}    \\  
BART & 0.0797 & -0.0709 & 0.0613 &  0.0021 &  -0.1115 \\
mBART & \tb{\increasebg{0.8849}}* & \tb{\increasebg{0.8170}}* & \tb{\increasebg{0.8638}}*  & \tb{\increasebg{0.8599}}*  & \tb{\increasebg{0.8725}}*   \\  \midrule
\multicolumn{6}{c}{\textbf{\textit{Coefficient of Determination (R$^2$) $\uparrow$}}} \\
T5 & -0.0909 & -0.1045 & -0.0750 &  -0.0942 & -0.1088 \\
mT5 &  \tb\smallincreasebg{0.6506}    &  \tb\smallincreasebg{0.5229} &  \tb\smallincreasebg{0.5994}     &  \tb\smallincreasebg{0.4697}      &  \tb\smallincreasebg{0.5810}   \\   
BART & -0.0637 & -0.1043 &  -0.0868 &  -0.0928 & -0.1116\\
mBART &   \tb{\increasebg{0.7776}} & \tb{\increasebg{0.6484}} & \tb{\increasebg{0.7369}}  & \tb{\increasebg{0.7361}}  & \tb{\increasebg{0.7546}}\\
\midrule
\multicolumn{6}{c}{\textbf{\textit{Root Mean Squared Error (RMSE)} $\downarrow$}} \\
T5 & 0.2218 & 0.2196 & 0.2180 & 0.2218 & 0.2219 \\
mT5 &  \tb\smallincreasebg{0.1271}     &   \tb\smallincreasebg{0.1443}     &  \tb\smallincreasebg{0.1331}     &   \tb\smallincreasebg{0.1544}       &  \tb\smallincreasebg{0.1364}    \\ 
BART & 0.2190 & 0.2195 & 0.2192  & 0.2217 & 0.2222\\
mBART & \tb{\increasebg{0.1002}} & \tb{\increasebg{0.1239}} & \tb{\increasebg{0.1079}} &  \tb{\increasebg{0.1089}} & \tb{\increasebg{0.1044}} \\
\bottomrule
\end{tabular}}
\caption{Regression results aligned with human society surveys. Statistically significant values with p $\leq$ 0.001 are marked with *. All correlations of multilingual models are positive and \colorbox{lightred}{outperform} monolingual.}
\vspace{-2mm}
\label{tb:reg_ret}
\end{table}

\begin{table*}[h]
\centering
\small
\renewcommand{\arraystretch}{.8}
\begin{tabular}{@{}l|lccccc||lccccc}
\toprule
Genre & Model & BLEU & R-1 & R-L & B-S & D-1 & Model & BLEU & R-1 & R-L & B-S & D-1  \\
\midrule \midrule
\multirow{2}{*}{Action} & mBART$_{zs}$  & 2.13 & 13.75  & 10.80  &  44.29  &  0.95 & mT5$_{zs}$ & 0.51 &  4.26  &  4.13  &    34.68  &  0.60  \\
                         & mBART$_{b}$  & \textbf{23.48} & \textbf{31.14}  & 28.87  &  54.37  &  \textbf{0.95}   & mT5$_b$ & 2.24 & 12.47   & 10.91   &  43.41  &  0.87  \\
                        & mBART$_{cul}$  & \decreasebg{23.44} & \decreasebg{30.98}  & \tb{\increasebg{29.08}} &  \tb{\increasebg{54.82}}  &  \decreasebg{0.94}  & mT5$_{cul}$ & \tb{\increasebg{2.41}} & \tb{\increasebg{12.62} } &  \tb{\increasebg{11.05}}  &   \tb{\increasebg{43.63}}  &  \tb{\increasebg{0.89}}  \\
                        \midrule
\multirow{2}{*}{Comedy}   & mBART$_{zs}$ & 2.34 & 14.09  & 11.16  &  44.18  &  0.93 & mT5$_{zs}$ & 0.55 &   4.62    &  4.48   &  34.80  &  0.58  \\
                        & mBART$_{b}$  & 2.60 & 13.56  & 11.52  &  42.47  &  0.85  & mT5$_b$ & 2.27 & 12.64  & 11.12  &  43.46  &  0.85 \\
                        & mBART$_{cul}$  & \tb{\increasebg{8.90}} & \tb{\increasebg{19.19}}  & \tb{\increasebg{16.67}}  &  \tb{\increasebg{46.40}}  &  \tb{\increasebg{0.93}}   & mT5$_{cul}$ &  \tb{\increasebg{2.68}} & \tb{\increasebg{13.22}}  &  \tb{\increasebg{11.50}}    &  \tb{\increasebg{43.99}}  &  \tb{\increasebg{0.90}} \\
                        \midrule
\multirow{2}{*}{Drama}  & mBART$_{zs}$  & 0.09 &  9.88  &  9.18  &  37.04  &  0.00   & mT5$_{zs}$ &  0.66 &   4.64  &  4.49  &  34.91  &  0.59 \\
                        & mBART$_{b}$  & 2.43 & \textbf{14.40}  & 11.30  &  \textbf{44.64}  &  \textbf{0.97}   & mT5$_b$ & 2.31 & 12.80  & 11.24  &  43.82  &  0.82 \\
                        & mBART$_{cul}$  & \tb{\increasebg{2.67}} & \decreasebg{13.95}  & \tb{\increasebg{12.02}}  &  \decreasebg{44.13}  &  \decreasebg{0.92} & mT5$_{cul}$ & \tb{\increasebg{2.53}} & \tb{\increasebg{13.01}}  & \tb{\increasebg{11.37}}   &  \tb{\increasebg{44.22}}  &  \tb{\increasebg{0.88}} \\
                        \midrule
\multirow{2}{*}{Romance}  & mBART$_{zs}$  & 2.26 & 14.25  & 11.24  &  44.29  &  \textbf{0.96}  & mT5$_{zs}$ & 0.58 &  4.67  &  4.53   &  34.74  &  0.58 \\
                        & mBART$_{b}$  & \textbf{14.91} & \textbf{24.03}  & \textbf{21.77}  &  49.64  &  0.95  & mT5$_b$ & \tb{2.28} &   \tb{12.95}    &   \tb{11.40}     &   \tb{43.97}    &   \tb{0.85}    \\
                        & mBART$_{cul}$  & \decreasebg{14.13} & \decreasebg{23.58}  & \decreasebg{21.23}  &  \tb{\increasebg{49.75}}  &  \decreasebg{0.95}  & mT5$_{cul}$ &   \decreasebg{2.17 } &   \decreasebg{12.66 }  &    \decreasebg{11.17 }   &    \decreasebg{43.78 }  &    \decreasebg{0.83 }   \\
                        \midrule
\multirow{2}{*}{Crime}  & mBART$_{zs}$  & 2.15 & 13.70  & 10.77  &  44.25  &  0.99  & mT5$_{zs}$ &  0.52 & 4.19  &  4.07   &  34.73  &  0.59  \\
                        & mBART$_{b}$  & 12.11 & 21.34  & 19.10  &  48.25  &  0.98   & mT5$_b$ & 2.14 & 12.10  & 10.58  &  43.28  &  0.85 \\
                        & mBART$_{cul}$  & \tb{\increasebg{12.95}} & \tb{\increasebg{22.07}}  & \tb{\increasebg{19.85}}  &  \tb{\increasebg{48.81}}  &  \tb{\increasebg{0.98}}  & mT5$_{cul}$ &  \tb{\increasebg{2.36}} & \tb{\increasebg{12.49}}  & \tb{\increasebg{10.92}}     &  \tb{\increasebg{43.57}}  &  \tb{\increasebg{0.89}} \\
\bottomrule
\end{tabular}
\caption{Prediction results for the multi-turn dialogue prediction task, demonstrating the impact of our proposed cultural enhancement on various genres. It reveals \colorbox{lightred}{improvements} in four genres, while one genre experienced a \colorbox{lightgreen}{decrease}. \#Avg is the average number of tokens by mT5 tokenizer. Vocab is the total vocabulary size. }
\label{tb:prediction_ret}
\end{table*}

\paragraph{Multi-Turn Dialogue Prediction.}
Table \ref{tb:prediction_ret} presents the results of our proposed cultural enhancement approach for multi-turn dialogue prediction. Pre-trained models without fine-tuning on cuDialog mBART$_{zs}$ and mT5$_{zs}$ exhibit weaker capabilities in dialogue prediction, resulting in lower values and shorter sentence length than fine-tuned models mBART$_{b}$ and mT5$_{b}$. This can be attributed to their pre-training tasks, which primarily focus on machine translation rather than dialogue generation. However, with cultural enhancement mBART$_{cul}$ and mT5$_{cul}$, dialogue prediction on most genres achieves better alignment with references and produces more diverse results, as evidenced by enhancements in both BLEU and Distinction metrics. Thus, it can be inferred that integrating cultural dimensions into dialogue agents leads to enhanced performance across various genres. Despite the improvements observed, there is still a need for further enhancement to improve the model's ability to comprehend and generate coherent responses in long-term dialogues, as supported by lower BLEU values consistent with prior work. 

Furthermore, we can find that the outcomes of mBART align consistently with that of mT5 model, which demonstrate enhanced metrics across the Action, Comedy, Drama, and Crime genres, except for Romance. Notably, improvements on mBART is more significant than mT5, which is consistent with the regression task in Table \ref{tb:reg_ret}. Our findings confirm the effectiveness of our cultural enhancement approach in improving dialogue prediction, aligning with references. To illustrate how the cultural attributes boost model performance, we provide the illustrative example of our generation results of mBART in Appendix \ref{ax:case_study}. 

\section{Discussion}
In our investigation regarding culture identification, we strive to explore the extent to which models can effectively capture cultural attributes within the context of cuDialog (\tb{RQ1}). Additionally, we examine the integration of these identified cultural attributes into the demanding task of multi-turn dialogue prediction, thereby yielding outcomes that are both more satisfactory and diverse. This empirical analysis provides compelling evidence that incorporating cultural considerations can improve the performance of dialogue agents, thus validating the notion that cultural awareness plays a crucial role in enhancing their effectiveness (\tb{RQ2}). 

\begin{figure}[t]
	\centering
	\includegraphics[width=1.0\columnwidth]{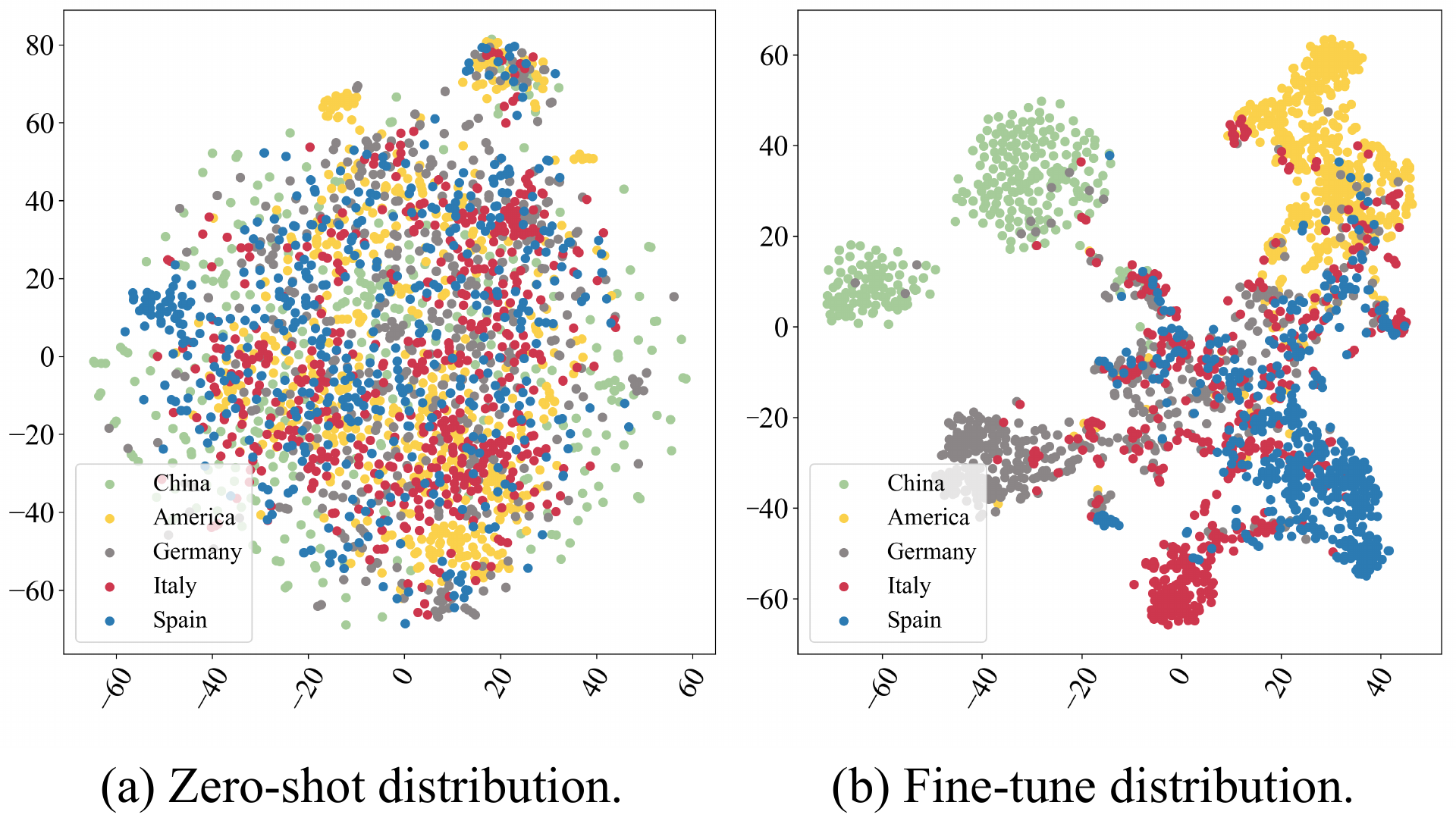}
	\caption{mT5 t-SNE before (left) and after (right) fine-tuning on regression. For clarity, we only select five countries as an example.}
	\label{fig:discussion_vec}
\end{figure}

\paragraph{Multilingual vs monolingual.} In cultural studies, the prevailing approach often focuses on languages associated with specific countries \cite{zhang2022mdia, kabra2023multi, keleg2023dlama}. However, we argue that models can acquire cultural attributes beyond linguistic distinctions alone. Capturing the essence of cultural phenomena, including values and beliefs, presents a complex challenge that requires empirical investigation \cite{hershcovich-etal-2022-challenges}. To validate our perspective, we randomly select 2,500 samples from five distinct cultures and visualize their representations using t-SNE based on the mT5 model. Figure \ref{fig:discussion_vec}(a) shows that zero-shot models struggle to differentiate between different cultural cases effectively. However, after fine-tuning the models with cuDialog, Figure \ref{fig:discussion_vec}(b) demonstrates a significant improvement in the separability of the representations. This indicates that incorporating cultural dimensions as guidance during fine-tuning facilitates the injection of implicit cultural features into language models.

\begin{figure}[t]
	\centering
	\includegraphics[width=0.45\textwidth]{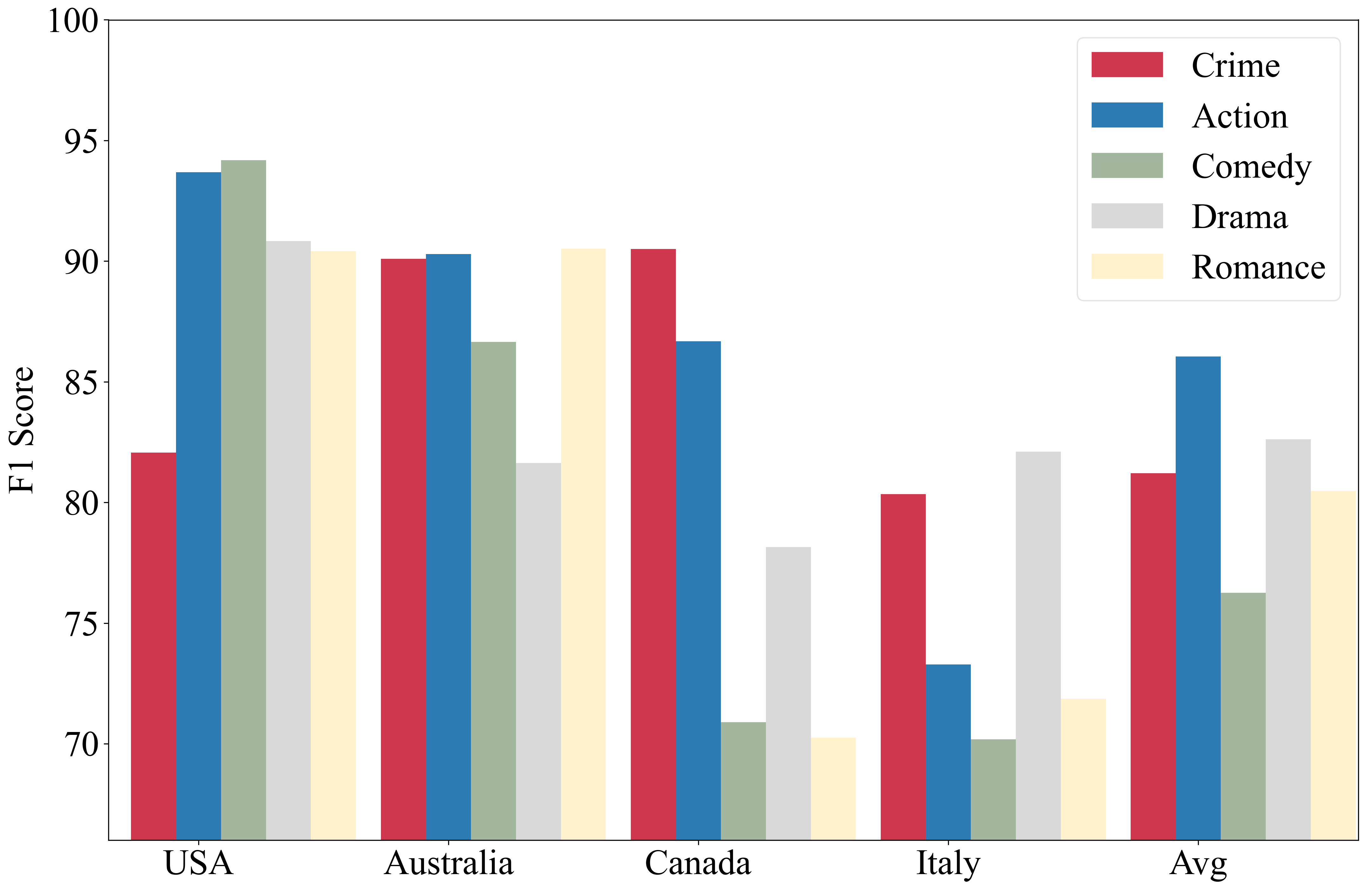}
	\caption{mBERT classification results, revealing clear distinctions in the classification capabilities of models across different cultures and genres.}
	\label{fig:classificaton_results_main}
\end{figure}

\paragraph{Cultural cues in cuDialog.}
Figure \ref{fig:classificaton_results_main} illustrates the significant variation in mBERT F1 scores for classification across cultures and genres. Notably, mBERT demonstrates a strong ability to identify American, Australian, and Canadian cultures, with particularly high performance in identifying American culture. These findings align with previous studies \cite{arora-etal-2023-probing, cao-etal-2023-assessing}. The dominance of the English training corpus \cite{ouyang2022training} contributes to a strong cultural embedding that may overshadow other cultures. Interestingly, \textit{Crime} and \textit{Action} dialogues consistently exhibit strong classification across all cultures, indicating a strong cultural component in these genres. This highlights the presence of cultural cues in cuDialog, resulting in distinct cultural representations.

\section{Conclusion}
We introduced cuDialog, utilizing OpenSubtitles 2018 for cultural identification and enhancing dialogue tasks. Our approach goes beyond dialogue texts by introducing culture classification and regression tasks, capturing both coarse-grained and fine-grained cultural knowledge. By leveraging cues from cultural value surveys, we bridge the cultural nuances between dialogue agents and human society, resulting in effective dialogue prediction adaptation. Further research in this area will advance the design of culturally aware dialogue systems that better meet user expectations.

\section*{Limitations}
\label{sec:Limitations and Future Work}
While our work has achieved good performance and shown promising results in enhancing dialogue tasks through incorporation of cultural cues, there are still limitations in our work.

The reference-based approach for multi-turn dialogue prediction evaluation is limited due to the subjectivity and variability of the task. A coherent and appropriate continuation may receive low scores simply because it diverges from the single reference in our dataset. 

The OpenSubtitles 2018 English corpus we used has inherent artifacts as it is a combination of human translations and machine-generated translations. Although we acknowledge that human translations tend to adapt to target cultures, we believe that distinct cultural differences can still be captured based on our observations.

Furthermore, we recognize that determining the cultural norm to align with remains an unresolved issue, as extensively discussed in \citet{gabriel2020artificial}. Our approach continues to be grounded in the premise that Chatbots should align to meet the needs of the majority of users, thereby aligning with individuals from diverse cultural backgrounds. 

We adopt human survey dimensions as cultural representations, despite its extensively aligned with human society, the intensity of the intervention is relatively soft. However, we believe that this study is still useful in highlighting the challenges of boosting the performance of dialogue agents by cultural considerations. In the future, we plan to explore the feasibility of collecting paired multicultural dialogues from conversation bots and utilizing structural cultural knowledge to guide the adaptation of cultural dialogues, which can be potential to provide further insights into incorporating cultural understanding into dialogue systems.

\section*{Ethics Statement}
Given the current gap in cross-cultural dialogue datasets within existing research, we have proposed constructing such datasets using existing dialogue corpora. However, obtaining paired cultural annotations for each dialogue presents a unique and open challenge, especially for benchmarking purposes. Ensuring the quality and accuracy of our multicultural dataset is crucial.

Our cultural dimension scores are derived from survey results obtained from a comprehensive sample of 117,000 matched employees across various countries, encompassing all the cultures of interest in our study.\footnote{\url{https://en.wikipedia.org/wiki/Hofstede\%27s_cultural_dimensions_theory}} Furthermore, in terms of genre labels, we utilize the annotations provided by OpenSubtitles, which are included in the original resource and annotated by its creators.
Our utilized datasets, including OpenSubtitles and Hofstede Cultural Survey,\footnote{Survey: \url{https://geerthofstede.com/research-and-vsm/vsm-2013}. Human society results: \url{https://geerthofstede.com/research-and-vsm/dimension-data-matrix/}} are publicly available and do not raise any privacy concerns. We have maintained the integrity of the data and adhered to privacy standards by not introducing any additional corpus or cultural annotations.
The OpenSubtitles is released with the GNU General Public License v3.0.\footnote{https://www.gnu.org/licenses/gpl-3.0.en.html} We will release our processed version with the same license.

We acknowledge that our analysis is based on the assumption that language accurately represents culture. However, we recognize that this notion may not be entirely congruent, as culture is complex, dynamic and highly diverse within countries and languages. This is especially true in cases where multiple official languages exist in a country, or where a language is spoken in multiple countries. Despite this limitation, our research still holds value as we identify a promising combination of existing corpora for our work.

Despite the above ethical considerations, this paper represents one of the initial endeavors in addressing cultural identification and cross-cultural dialogue enhancement, making it a pioneering effort in exploring the cultural adaptability of dialogue agents. We believe this research direction has the potential to mitigate cultural biases and facilitate honest, respectful and informative cross-cultural communication between humans, with the assistance of AI.

\section{Acknowledgement}
Thanks to the anonymous reviewers for their helpful feedback. The authors gratefully acknowledge financial support from China Scholarship Council. (CSC No. 202206160052).

\bibliography{anthology,custom}
\bibliographystyle{acl_natbib}

\appendix
\section{Hofstede Cultural Survey}
\label{ax:hosfeted_survey}

This survey is one of the most commonly used cross-cultural tools developed by Dutch social psychologist, Geert Hofstede, aiming to measure cultural distinctions among countries. Six cultural dimensions are proposed by this survey, including:
\begin{itemize}
    \item \textbf{Power Distance (pdi).} It measures the acceptance of unequal power distribution within organizations and institutions.
    \item \textbf{Individualism (idv).} It explores the extent to which individuals are integrated into groups.
    \item \textbf{Uncertainty Avoidance (uai).} It assesses the individuals' attitude to something unexpected, unknown, or away from the status quo.
    \item \textbf{Masculinity (mas).} It measures individuals' preference in society for achievement, heroism, assertiveness, and material rewards for success.
    \item \textbf{Long-term Orientation (lto).} It measures the focus on traditions and steadfastness (short-term) versus adaptability and pragmatic problem-solving (long-term).
    \item \textbf{Indulgence (ivr).} It measures the degree of societal norms in allowing individuals to freely fulfill their desires.
\end{itemize}
This survey will ask participants to answer 24 questions and drive each dimension scores $S_i$ based on four related questions $\mathcal{Q}_i$ by:
\begin{equation}
    S_i = \lambda^0_i (\mathcal{Q}^0_i - \mathcal{Q}^1_i) + \lambda^1_i (\mathcal{Q}^2_i - \mathcal{Q}^3_i) + \mathcal{C}_i
\label{eq:score_cal}
\end{equation}
where $\lambda_i$ is the hyper-parameter and $\mathcal{C}_i$ is a constant. Detailed values for $\lambda_i$ and $\mathcal{Q}_i$ are listed in Table \ref{tb:hyper_params_survey}. The results of our used cultures are listed in Table \ref{tb:hosfted_statis}. Besides, given Hofstede scores, we tabulated all the cases in our proposed cuDialog in Table \ref{tb:extra_statis_cuDialog}.

\begin{table}[]
\centering
\resizebox{0.98\columnwidth}{!}{
\begin{tabular}{c|cl}
\toprule
\textbf{Dimension} & \textbf{Coefficient} $\lambda_i$ & \textbf{Questions} $\mathcal{Q}_i$ \\ \midrule \midrule
   pdi    &  35, 25     &   7, 2, 20,23   \\
   idv       &    35, 35         &  4, 1, 9, 6     \\
   mas       &    35, 35         &  5, 3, 8, 10     \\
   uai       &    40, 25         &  18, 15, 21, 24  \\
   lto       &    40, 25         &  13, 14, 19, 22  \\
   ivr       &    35, 40         &  12, 11, 17, 16  \\ \bottomrule
\end{tabular}}
\caption{\label{tb:hyper_params_survey} The hyper-parameter setting of six cultural dimension metrics in the Hofstede Culture Survey.}
\end{table}

\begin{table}[t]
\centering
\resizebox{0.45\textwidth}{!}{
\begin{tabular}{l|cccccc}
\toprule
\multirow{2}{*}{\textbf{Cul}} & \multicolumn{6}{c}{\textbf{Cultural Dimension}}                                      \\ \cline{2-7} 
                         & \cellcenter{\textbf{pdi}} & \cellcenter{\textbf{idv}} & \cellcenter{\textbf{uai}} & \cellcenter{\textbf{mas}} & \cellcenter{\textbf{lto}} & \cellcenter{\textbf{ivr}} \\ \midrule\midrule
US  & 40.0 & 91.0 & 62.0 & 46.0 & 26.0 & 68.0  \\
UK  & 35.0 & 89.0 & 66.0 & 35.0 & 51.0 & 69.0  \\
FR  & 68.0 & 71.0 & 43.0 & 86.0 & 63.0 & 48.0 \\
JA  & 54.0 & 46.0 & 95.0 & 92.0 & 88.0 & 42.0 \\
GM  &  35.0 & 67.0 & 66.0 & 65.0 & 83.0 & 40.0  \\
CA  & 39.0 & 80.0 & 52.0 & 48.0 & 36.0 & 68.0  \\
IT  &  50.0 & 76.0 & 70.0 & 75.0 & 61.0 & 30.0  \\
KS  & 60.0 & 18.0 & 39.0 & 85.0 & 100.0 & 29.0 \\
IN  & 77.0 & 48.0 & 56.0 & 40.0 & 51.0 & 26.0  \\
SP  & 57.0 & 51.0 & 42.0 & 86.0 & 48.0 & 44.0 \\
AS  &  38.0 & 90.0 & 61.0 & 51.0 & 21.0 & 71.0 \\
CH  & 80.0 & 20.0 & 66.0 & 30.0 & 87.0 & 24.0 \\
SE  & 31.0 & 71.0 & 5.0 & 29.0 & 53.0 & 78.0 \\   \bottomrule
\end{tabular}}
\caption{Statistical results of cultural indicators of the human society survey. }
\label{tb:hosfted_statis}
\end{table}

\section{Significance Check}

To ascertain the non-trivial nature of our experimental findings,  we pass our experiment results of multi-turn dialogue prediction task through a statistical significance test, aiming to show the effectiveness of our improvements. To achieve this, we have employed a widely recognized tool as outlined in \citet{dror-etal-2018-hitchhikers} and \citet{P18-1128}. Specifically, we format our predictions of each case and baseline's as required by \citet{P18-1128} and then conduct Anderson-Darling (ad) with the desirable significance level (alpha=0.05) and t-test. By comparing the BLEU metrics derived from the aforementioned mBART generation table, we have obtained the results presented in Table \ref{tb:ax_significant_check} (Yes denotes significant, Not denotes not significant). Notably, a substantial portion of the BLEU scores exhibit statistical significance when compared to the baseline outcomes.

\begin{table}[htbp]
\centering
\resizebox{0.48\textwidth}{!}{
\begin{tabular}{l|ccc|ccc}
\toprule
\multirow{2}{*}{Genre} & \multicolumn{3}{c|}{Anderson-Darling} & \multicolumn{3}{c}{t-test} \\
                       & BLEU-1     & BLEU-2     & BLEU-4     & BLEU-1  & BLEU-2  & BLEU-4 \\ \midrule \midrule
Comedy                 & $\checkmark$ & $\checkmark$ & $\checkmark$ & $\times$ & $\checkmark$ & $\checkmark$       \\
Drama                  & $\checkmark$ & $\checkmark$ & $\checkmark$ & $\checkmark$ & $\checkmark$ & $\checkmark$         \\
Romance                & $\checkmark$ & $\checkmark$ & $\checkmark$ & $\checkmark$ & $\checkmark$ & $\checkmark$           \\
Crime                 & $\checkmark$ & $\checkmark$ & $\checkmark$ & $\checkmark$ & $\checkmark$ & $\checkmark$           \\
Action                 & $\checkmark$ & $\checkmark$ & $\checkmark$ & $\checkmark$ & $\checkmark$ & $\checkmark$      \\    \bottomrule
\end{tabular}}
\caption{Statistical significance test for multi-turn dialogue prediction task in mBART model, where $\checkmark$ denotes significant and $\times$ denotes not significant. }
\label{tb:ax_significant_check}
\end{table}

\section{Case Study}
\label{ax:case_study}
To illustrate how the cultural attributes boost model performance, we present an illustrative example within the Crime genre test set below, which is from an Italian film titled \textit{"Pasolini Un Delitto Italiano - Marco Tullio Giordana (1995)"}, as is shown in Table \ref{tb:ax_cast_study}.

In this instance, it is evident that the history sentences are talking about politics and crimes and the attitude of golden sentences is ``refusing to tell the truth'', whose topic and attitude aligns closely with our generated sentences, but quite different from the baseline model’s output. Moreover, for the Romance genre, studies show that romance can indeed deviate from accepted social and cultural norms, agreeing with our obtained results. For example, Asian romance movies often challenge values such as obeying authority, adherence to cultural norms and putting society before self \cite{rahman2013life}.

\begin{table}
\centering
\begin{tabularx}{\columnwidth}{X}
\toprule
\small \textbf{History:}
\textit{...there are crimes that have impunity. There is such hostility around the victim. It is able to escape unpunished. Then, there is no doubt and there is a plot of power, to silence the voice of an opponent.} \\
\small
\textbf{Golden Predictions:}
\textit{I do not have any truth. You are the ones who have any doubts…} \\
\small
\textbf{Baseline Prediction:} I am sorry, but I do not understand what you are saying. You are asking me to silence the voice of an opponent? \\ \midrule
\small
\textbf{Ours Prediction: } I am not a communist. I am a pacifist. I will not talk about hatred or hatred in public. \\
\bottomrule
\end{tabularx}
\caption{Case study for multi-turn dialogue prediction.}
\label{tb:ax_cast_study}
\end{table}

\section{Hyper Parameter Setting}
\label{ax:hyper_parameters}
To facilitate the reproducibility of our training process for culture classification, culture regression, and multi-turn dialogue prediction tasks, we provide a comprehensive list of the hyper-parameters used to achieve the best results on our proposed datasets, as demonstrated in Table \ref{tb:para_setting}.

\begin{table}[htbp]
\centering
\resizebox{0.45\textwidth}{!}{
\begin{tabular}{l|c|c|c}
\toprule
\textbf{Parameter}   & Classification & Regression & Prediction \\
\midrule
\emph{Learning rate} & $3e^{-5}$ & $1e^{-4}$ & $1e^{-4}$ \\
\emph{Batch size}    & 128        & 128 & 64   \\
\emph{Epochs}  		  &  30    & 30 & 50    \\
\emph{Num Labels} & 13 & 6 & - \\
\emph{GPU Num} & 1 & 1 & 2 \\
\emph{Warmup Steps} & 156 & 0 & 0 \\
\emph{Early Stop} & \checkmark &  \checkmark & \checkmark \\
\emph{Early Stop Patience} & 3 & 2 & 2 \\ 
\emph{Repetition Penalty} & - & - & 1.2 \\
\emph{Num Beams} & - & - & 3 \\
\bottomrule
\end{tabular}}
\caption{\label{tb:para_setting} The hyper-parameter settings of the best results on our proposed three tasks.}
\end{table}

\begin{table*}[ht]
\centering
\begin{tabular}{l|ccccc}
\toprule
\multirow{2}{*}{\textbf{Culture}} & \multicolumn{5}{c}{\textbf{Topics}}                                      \\ \cline{2-6} 
                         & \cellcenter{\textbf{Action}} & \cellcenter{\textbf{Comedy}} & \cellcenter{\textbf{Drama}} & \cellcenter{\textbf{Romance}} & \cellcenter{\textbf{Crime}} \\ \midrule\midrule
USA(US)  & 15,221 & 15,110 & 11,820 & 14,081 & 11,154  \\
Britain (UK)  & 11,233 & 16,076 & 11,260 & 10,336 & 11,771  \\
France (FR) & 10,598 & 12,953 & 8,771 & 9,403 & 12,021  \\
Japan (JA)  & 7,601 & 11,097 & 8,695 & 7,778 & 8,311  \\
Germany (GM)  & 10,163 & 12,459 & 11,009 & 10,106 & 11,169  \\
Canada (CA)  & 10,171 & 13,795 & 9,010 & 11,269 & 10,004  \\
Italy (IT)  & 8,873 & 17,378 & 15,890 & 11,810 & 13,056  \\
South Korea (KS)  & 7,128 & 7,487 & 9,070 & 8,787 & 9,349  \\
India (IN) & 13,783 & 16,164 & 14,268 & 15,407 & 13,278  \\
Spain (SP) & 10,350 & 13,861 & 9,833 & 10,029 & 12,180  \\
Australia (AS)  & 12,107 & 12,953 & 10,114 & 14,117 & 10,872  \\
China (CH)  & 11,202 & 12,020 & 10,751 & 10,262 & 11,111  \\
Sweden (SZ)  & 8,648 & 11,696 & 8,478 & 10,484 & 9,585  \\
\bottomrule
\end{tabular}
\caption{Detailed statistics of cuDialog, consisting of 13 cultural backgrounds and 5 conversation genres. The dataset includes movie subtitles between individuals from different cultures discussing various genres such as comedy, romance, etc. The 13 cultural backgrounds represented in the dataset include but are not limited to American, Chinese, Indian, and Japanese cultures.}
\label{tb:extra_statis_cuDialog}
\end{table*}

\begin{figure*}[t]
	\centering
	\includegraphics[width=1.0\textwidth]{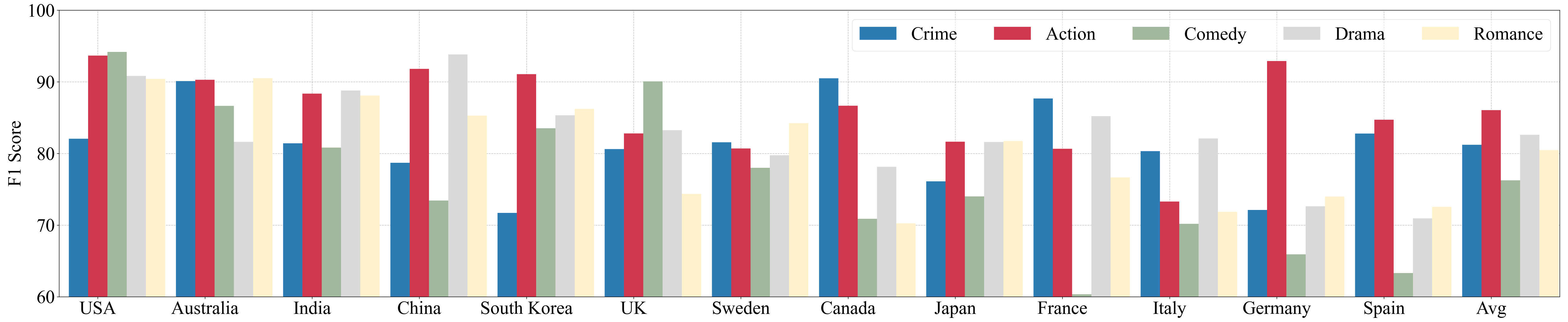}
	\caption{mBERT classification results, showing cultural features vary among countries and genres. }
	\label{fig:classificaton_results}
\end{figure*}

\section{Pre-trained Models Download}
\label{ax:pre_trained_models}
All BERT pre-trainied models adopted in Table \ref{tb:classificaton_results}, \ref{tb:reg_ret} and \ref{tb:prediction_ret} are published by \citep{wolf-etal-2020-transformers}.
In order to help reproduce our work and use our code easily, we summarize the download links of the pre-trained models as follows.

\para{Culture Classification.}
    \begin{itemize}
      \item BERT \\
      \url{https://huggingface.co/bert-base-uncased}
      \item multilingual BERT \\
      \url{https://huggingface.co/bert-base-multilingual-cased}
      \item RoBERTa \\
      \url{https://huggingface.co/roberta-base}
      \item XLM-RoBERTa \\
      \url{https://huggingface.co/xlm-roberta-base}
    \end{itemize}

\para{Culture Regression \& Dialogue Prediction.}
    \begin{itemize}
      \item T5 \\
      \url{https://huggingface.co/t5-base}
      \item mT5 \\
      \url{https://huggingface.co/mt5-base}
      \item BART \\
      \url{https://huggingface.co/facebook/bart-base}
      \item mBART \\
      \url{https://huggingface.co/facebook/mbart-large-50}
    \end{itemize}

\section{Classification Results}
\label{ax:classification_ret}

The results of the culture classification task, including recall, precision, and F1 scores, are presented here, including BERT (Table \ref{tb:ax_bert_cls_ret}), mBERT (Table \ref{tb:ax_mbert_cls_ret}), Roberta (Table \ref{tb:ax_roberta_cls_ret}), and mRoberta (Table \ref{tb:ax_mroberta_cls_ret}). Additionally, for enhanced clarity and visual representation, we offer a comprehensive comparison of F1 scores for all cultures and topics of mBERT in Figure \ref{fig:classificaton_results}, with the complete version depicted in Figure \ref{fig:classificaton_results_main}. These findings provide valuable insights into the performance and effectiveness of different models in accurately classifying cultures and topics, contributing to the advancement of the field.

\begin{table*}[]
\centering
\resizebox{\textwidth}{!}{
\begin{tabular}{l|ccc|ccc|ccc|ccc|ccc}
\toprule
\multirow{2}{*}{\textbf{Cul}} & \multicolumn{3}{c|}{\textbf{Action}}  & \multicolumn{3}{c|}{\textbf{Comedy}}     & \multicolumn{3}{c|}{\textbf{Drama}} & \multicolumn{3}{c|}{\textbf{Romance}} & \multicolumn{3}{c}{\textbf{Crime}} \\ \cline{2-16} 
  & \cellcenter{\textbf{Rec}} & \cellcenter{\textbf{Pre}} & \cellcenter{\textbf{F1}} 
  & \cellcenter{\textbf{Rec}} & \cellcenter{\textbf{Pre}} & \cellcenter{\textbf{F1}}
  & \cellcenter{\textbf{Rec}} & \cellcenter{\textbf{Pre}} & \cellcenter{\textbf{F1}}
  & \cellcenter{\textbf{Rec}} & \cellcenter{\textbf{Pre}} & \cellcenter{\textbf{F1}}
  & \cellcenter{\textbf{Rec}} & \cellcenter{\textbf{Pre}} & \cellcenter{\textbf{F1}}
\\ \midrule\midrule
US & 96.03 & 92.82 & 94.40  & 92.66 & 92.71 & 92.68 & 96.59 & 90.96 & 93.69 & 94.32 & 90.08 & 92.15 & 85.70  & 95.65 & 90.40  \\
UK & 77.42 & 94.81 & 85.24 & 91.11 & 92.53 & 91.82 & 81.12 & 92.55 & 86.46 & 75.51 & 82.96 & 79.06 & 90.13 & 84.39 & 87.17 \\
FR & 84.18 & 81.96 & 83.06 & 56.19 & 60.75 & 58.38 & 96.26 & 74.79 & 84.18 & 84.24 & 79.24 & 81.66 & 93.76 & 83.64 & 88.41 \\
JA & 79.78 & 91.73 & 85.34 & 74.34 & 79.38 & 76.78 & 77.66 & 85.49 & 81.39 & 73.80  & 95.43 & 83.24 & 84.59 & 76.36 & 80.27 \\
GM & 93.80  & 89.10  & 91.39 & 59.52 & 82.64 & 69.20  & 67.40  & 81.16 & 73.64 & 73.57 & 75.17 & 74.36 & 73.28 & 83.05 & 77.86 \\
CA & 83.44 & 92.53 & 87.75 & 88.78 & 65.82 & 75.60  & 86.53 & 73.47 & 79.47 & 79.43 & 68.48 & 73.55 & 94.48 & 89.68 & 92.01 \\
IT & 81.69 & 71.87 & 76.47 & 63.64 & 77.00    & 69.69 & 82.72 & 81.45 & 82.08 & 70.95 & 76.36 & 73.56 & 88.43 & 83.07 & 85.67 \\
KS & 93.71 & 94.12 & 93.92 & 93.69 & 75.25 & 83.46 & 85.84 & 84.39 & 85.11 & 88.69 & 94.24 & 91.38 & 71.84 & 79.57 & 75.51 \\
IN & 90.07 & 94.59 & 92.28 & 84.52 & 86.58 & 85.54 & 89.91 & 91.30  & 90.60  & 84.91 & 93.77 & 89.12 & 81.52 & 94.47 & 87.52 \\
SP & 95.38 & 84.94 & 89.86 & 57.68 & 64.06 & 60.70  & 66.89 & 70.18 & 68.49 & 75.22 & 75.42 & 75.32 & 89.41 & 82.25 & 85.68 \\
AS & 95.78 & 92.05 & 93.88 & 91.49 & 75.38 & 82.66 & 85.76 & 89.25 & 87.47 & 96.72 & 85.74 & 90.90  & 92.68 & 84.47 & 88.38 \\
CH & 96.11 & 92.39 & 94.21 & 77.17 & 73.84 & 75.47 & 95.72 & 92.61 & 94.14 & 85.48 & 82.24 & 83.83 & 85.58 & 87.14 & 86.35 \\
SE & 84.10 & 81.10 & 82.57 & 77.50  & 75.48 & 76.48 & 80.49 & 84.26 & 82.33 & 88.87 & 82.71 & 85.68 & 84.40  & 93.77 & 88.84 \\ \midrule
\textbf{AVG} &
  \textbf{88.58} &
  \textbf{88.77} &
  \textbf{88.49} &
  \textbf{77.56} &
  \textbf{77.03} &
  \textbf{76.80} &
  \textbf{84.07} &
  \textbf{83.99} &
  \textbf{83.77} &
  \textbf{82.44} &
  \textbf{83.22} &
  \textbf{82.60} &
  \textbf{85.83} &
  \textbf{85.96} &
  \textbf{85.70} \\ 
\bottomrule
\end{tabular}}
\caption{Recall (Rec), Precision(Pre) and F1 Performance of Dialogue Culture Classification Model based on BERT. The performance indicators are reported for 13 different cultural categories.}
\label{tb:ax_bert_cls_ret}
\end{table*}

\begin{table*}[]
\centering
\resizebox{\textwidth}{!}{
\begin{tabular}{l|ccc|ccc|ccc|ccc|ccc}
\toprule
\multirow{2}{*}{\textbf{Cul}} & \multicolumn{3}{c|}{\textbf{Action}}  & \multicolumn{3}{c|}{\textbf{Comedy}}     & \multicolumn{3}{c|}{\textbf{Drama}} & \multicolumn{3}{c|}{\textbf{Romance}} & \multicolumn{3}{c}{\textbf{Crime}} \\ \cline{2-16} 
  & \cellcenter{\textbf{Rec}} & \cellcenter{\textbf{Pre}} & \cellcenter{\textbf{F1}} 
  & \cellcenter{\textbf{Rec}} & \cellcenter{\textbf{Pre}} & \cellcenter{\textbf{F1}}
  & \cellcenter{\textbf{Rec}} & \cellcenter{\textbf{Pre}} & \cellcenter{\textbf{F1}}
  & \cellcenter{\textbf{Rec}} & \cellcenter{\textbf{Pre}} & \cellcenter{\textbf{F1}}
  & \cellcenter{\textbf{Rec}} & \cellcenter{\textbf{Pre}} & \cellcenter{\textbf{F1}}
\\ \midrule\midrule
US & 92.06 & 95.36 & 93.68 & 91.29 & 97.28 & 94.19 & 96.04 & 86.15 & 90.83 & 90.41 & 90.41 & 90.41 & 78.03 & 86.54 & 82.07 \\
UK & 71.67 & 98.03 & 82.80 & 90.94 & 89.22 & 90.07 & 79.61 & 87.25 & 83.26 & 70.13 & 79.13 & 74.36 & 86.65 & 75.40 & 80.63 \\
FR & 80.16 & 81.16 & 80.66 & 61.47 & 59.25 & 60.34 & 95.38 & 77.00 & 85.21 & 87.59 & 68.17 & 76.67 & 90.32 & 85.19 & 87.68 \\
JA & 75.51 & 88.89 & 81.65 & 70.11 & 78.40 & 74.02 & 77.76 & 85.86 & 81.61 & 75.40 & 89.22 & 81.73 & 79.85 & 72.72 & 76.12 \\
GM & 92.86 & 92.96 & 92.91 & 61.26 & 71.36 & 65.92 & 66.26 & 80.38 & 72.64 & 67.27 & 82.20 & 73.99 & 68.67 & 75.96 & 72.13 \\
CA & 85.30 & 88.11 & 86.68 & 90.30 & 58.35 & 70.89 & 80.64 & 75.81 & 78.15 & 68.80 & 71.76 & 70.25 & 93.20 & 87.97 & 90.51 \\
IT & 72.39 & 74.22 & 73.29 & 64.24 & 77.32 & 70.18 & 79.31 & 85.11 & 82.11 & 72.20 & 71.53 & 71.86 & 83.49 & 77.41 & 80.34 \\
KS & 94.15 & 88.21 & 91.08 & 83.84 & 83.21 & 83.52 & 86.60 & 84.12 & 85.34 & 92.10 & 81.07 & 86.23 & 65.92 & 78.59 & 71.70 \\
IN & 88.54 & 88.16 & 88.35 & 84.38 & 77.55 & 80.82 & 86.61 & 91.09 & 88.80 & 84.44 & 92.06 & 88.09 & 72.40 & 93.04 & 81.43 \\
SP & 91.54 & 78.83 & 84.71 & 56.27 & 72.36 & 63.31 & 65.55 & 77.32 & 70.95 & 77.17 & 68.45 & 72.55 & 88.62 & 77.69 & 82.79 \\
AS & 97.66 & 83.96 & 90.29 & 89.47 & 84.01 & 86.65 & 90.92 & 74.07 & 81.63 & 92.01 & 89.08 & 90.52 & 91.11 & 89.11 & 90.10 \\
CH & 96.59 & 87.51 & 91.82 & 75.00 & 71.94 & 73.44 & 95.49 & 92.25 & 93.84 & 82.21 & 88.60 & 85.29 & 78.18 & 79.24 & 78.71 \\
SE & 81.97 & 79.48 & 80.71 & 79.25 & 76.79 & 78.00 & 82.37 & 77.29 & 79.75 & 87.86 & 80.93 & 84.25 & 83.36 & 79.85 & 81.57 \\ \midrule
\textbf{AVG} &
  \multicolumn{1}{l}{\textbf{86.18}} &
  \textbf{86.53} &
  \textbf{86.05} &
  \multicolumn{1}{l}{\textbf{76.76}} &
  \textbf{76.70} &
  \textbf{76.26} &
  \multicolumn{1}{l}{\textbf{83.27}} &
  \textbf{82.59} &
  \textbf{82.62} &
  \multicolumn{1}{l}{\textbf{80.58}} &
  \textbf{80.97} &
  \textbf{80.48} &
  \multicolumn{1}{l}{\textbf{81.52}} &
  \textbf{81.44} &
  \textbf{81.21}\\  
\bottomrule
\end{tabular}}
\caption{Recall (Rec), Precision(Pre) and F1 Performance of Dialogue Culture Classification Model based on mBERT. The performance indicators are reported for 13 different cultural categories.}
\label{tb:ax_mbert_cls_ret}
\end{table*}

\begin{table*}[]
\centering
\resizebox{\textwidth}{!}{
\begin{tabular}{l|ccc|ccc|ccc|ccc|ccc}
\toprule
\multirow{2}{*}{\textbf{Cul}} & \multicolumn{3}{c|}{\textbf{Action}}  & \multicolumn{3}{c|}{\textbf{Comedy}}     & \multicolumn{3}{c|}{\textbf{Drama}} & \multicolumn{3}{c|}{\textbf{Romance}} & \multicolumn{3}{c}{\textbf{Crime}} \\ \cline{2-16} 
  & \cellcenter{\textbf{Rec}} & \cellcenter{\textbf{Pre}} & \cellcenter{\textbf{F1}} 
  & \cellcenter{\textbf{Rec}} & \cellcenter{\textbf{Pre}} & \cellcenter{\textbf{F1}}
  & \cellcenter{\textbf{Rec}} & \cellcenter{\textbf{Pre}} & \cellcenter{\textbf{F1}}
  & \cellcenter{\textbf{Rec}} & \cellcenter{\textbf{Pre}} & \cellcenter{\textbf{F1}}
  & \cellcenter{\textbf{Rec}} & \cellcenter{\textbf{Pre}} & \cellcenter{\textbf{F1}}
\\ \midrule\midrule
US & 95.20 & 92.36 & 93.76 & 89.50 & 94.16 & 91.77 & 95.97 & 86.90 & 91.21 & 91.29 & 93.38 & 92.32 & 77.27 & 97.58 & 86.24 \\
UK & 77.58 & 94.03 & 85.01 & 89.80 & 84.61 & 87.13 & 80.69 & 87.13 & 83.79 & 77.36 & 81.75 & 79.49 & 94.28 & 83.48 & 88.55 \\
FR & 82.38 & 79.01 & 80.66 & 59.94 & 54.76 & 57.24 & 94.83 & 77.59 & 85.35 & 84.14 & 84.98 & 84.55 & 96.48 & 84.35 & 90.01 \\
JA & 76.85 & 92.93 & 84.13 & 67.55 & 77.06 & 71.99 & 82.24 & 77.33 & 79.71 & 80.52 & 83.97 & 82.21 & 85.56 & 72.58 & 78.54 \\
GM & 92.66 & 94.22 & 93.43 & 56.04 & 70.52 & 62.45 & 55.24 & 85.75 & 67.20 & 72.34 & 81.10 & 76.47 & 71.59 & 78.13 & 74.72 \\
CA & 85.65 & 86.34 & 85.99 & 84.08 & 63.48 & 72.34 & 82.88 & 70.60 & 76.25 & 78.28 & 68.74 & 73.20 & 96.60 & 89.25 & 92.78 \\
IT & 80.08 & 75.36 & 77.64 & 71.36 & 69.09 & 70.20 & 86.47 & 75.29 & 80.50 & 69.39 & 85.26 & 76.51 & 89.01 & 81.57 & 85.13 \\
KS & 93.82 & 93.10 & 93.46 & 91.41 & 87.65 & 89.49 & 82.79 & 90.45 & 86.45 & 93.57 & 85.69 & 89.46 & 75.44 & 83.37 & 79.20 \\
IN & 92.16 & 93.65 & 92.90 & 84.72 & 78.06 & 81.26 & 89.20 & 95.13 & 92.07 & 85.79 & 92.88 & 89.19 & 73.22 & 97.42 & 83.60 \\
SP & 87.81 & 89.75 & 88.77 & 50.42 & 79.93 & 61.84 & 62.46 & 78.63 & 69.62 & 78.76 & 69.31 & 73.74 & 89.57 & 82.52 & 85.90 \\
AS & 94.77 & 91.48 & 93.10 & 86.86 & 79.17 & 82.83 & 89.96 & 79.61 & 84.47 & 95.89 & 91.49 & 93.64 & 94.69 & 93.57 & 94.13 \\
CH & 97.00 & 87.93 & 92.24 & 80.25 & 74.59 & 77.32 & 95.34 & 90.04 & 92.62 & 86.83 & 82.92 & 84.83 & 85.58 & 86.68 & 86.12 \\
SE & 86.00 & 78.29 & 81.96 & 76.43 & 73.14 & 74.75 & 76.62 & 87.75 & 81.81 & 89.97 & 82.39 & 86.02 & 85.86 & 81.91 & 83.84 \\ \midrule
\textbf{AVG} &
  \textbf{87.84} &
  \textbf{88.34} &
  \textbf{87.93} &
  \textbf{76.03} &
  \textbf{75.86} &
  \textbf{75.43} &
  \textbf{82.67} &
  \textbf{83.25} &
  \textbf{82.39} &
  \textbf{83.39} &
  \textbf{83.37} &
  \textbf{83.20} &
  \textbf{85.78} &
  \textbf{85.57} &
  \textbf{85.29} \\ \midrule
\bottomrule
\end{tabular}}
\caption{Recall (Rec), Precision(Pre) and F1 Performance of Dialogue Culture Classification Model based on Roberta. The performance indicators are reported for 13 different cultural categories.}
\label{tb:ax_roberta_cls_ret}
\end{table*}

\begin{table*}[]
\centering
\resizebox{\textwidth}{!}{
\begin{tabular}{l|ccc|ccc|ccc|ccc|ccc}
\toprule
\multirow{2}{*}{\textbf{Cul}} & \multicolumn{3}{c|}{\textbf{Action}}  & \multicolumn{3}{c|}{\textbf{Comedy}}     & \multicolumn{3}{c|}{\textbf{Drama}} & \multicolumn{3}{c|}{\textbf{Romance}} & \multicolumn{3}{c}{\textbf{Crime}} \\ \cline{2-16} 
  & \cellcenter{\textbf{Rec}} & \cellcenter{\textbf{Pre}} & \cellcenter{\textbf{F1}} 
  & \cellcenter{\textbf{Rec}} & \cellcenter{\textbf{Pre}} & \cellcenter{\textbf{F1}}
  & \cellcenter{\textbf{Rec}} & \cellcenter{\textbf{Pre}} & \cellcenter{\textbf{F1}}
  & \cellcenter{\textbf{Rec}} & \cellcenter{\textbf{Pre}} & \cellcenter{\textbf{F1}}
  & \cellcenter{\textbf{Rec}} & \cellcenter{\textbf{Pre}} & \cellcenter{\textbf{F1}}
\\ \midrule\midrule
US & 93.09 & 94.60  & 93.84 & 87.27 & 97.63 & 92.16 & 95.41 & 89.32 & 92.27 & 85.45 & 91.92 & 88.57 & 81.23 & 93.39 & 86.89 \\
UK & 70.53 & 97.28 & 81.77 & 91.20  & 87.26 & 89.19 & 79.76 & 84.42 & 82.02 & 68.37 & 74.47 & 71.29 & 88.72 & 80.39 & 84.35 \\
FR & 81.72 & 75.30  & 78.38 & 71.77 & 46.49 & 56.42 & 94.61 & 76.31 & 84.48 & 79.70  & 79.31 & 79.51 & 94.28 & 82.37 & 87.92 \\
JA & 80.45 & 86.06 & 83.16 & 78.66 & 63.58 & 70.32 & 75.61 & 86.25 & 80.58 & 74.94 & 87.04 & 80.54 & 76.72 & 81.75 & 79.16 \\
GM & 93.38 & 91.12 & 92.24 & 55.96 & 69.26 & 61.90  & 57.60  & 82.69 & 67.90  & 65.55 & 79.07 & 71.68 & 68.96 & 79.24 & 73.74 \\
CA & 85.12 & 88.90  & 86.97 & 86.73 & 60.66 & 71.39 & 77.27 & 61.50  & 68.49 & 70.33 & 67.62 & 68.95 & 93.54 & 89.88 & 91.67 \\
IT & 77.13 & 75.56 & 76.34 & 61.12 & 78.65 & 68.78 & 84.28 & 77.59 & 80.80  & 70.48 & 76.05 & 73.16 & 88.78 & 76.12 & 81.96 \\
KS & 95.03 & 86.62 & 90.63 & 87.37 & 84.60  & 85.96 & 86.69 & 83.75 & 85.19 & 93.66 & 76.27 & 84.08 & 76.41 & 78.31 & 77.35 \\
IN & 90.44 & 88.38 & 89.40  & 87.04 & 80.45 & 83.62 & 87.02 & 92.51 & 89.68 & 86.26 & 88.81 & 87.52 & 78.75 & 93.57 & 85.52 \\
SP & 89.68 & 82.46 & 85.92 & 53.57 & 74.96 & 62.49 & 58.95 & 75.56 & 66.23 & 70.97 & 72.91 & 71.93 & 86.96 & 86.96 & 86.96 \\
AS & 93.36 & 91.30  & 92.32 & 87.75 & 79.93 & 83.66 & 89.10  & 76.39 & 82.26 & 93.10  & 87.59 & 90.26 & 92.38 & 90.68 & 91.52 \\
CH & 94.68 & 89.66 & 92.10  & 73.25 & 83.63 & 78.10  & 96.26 & 85.08 & 90.32 & 82.50  & 83.30  & 82.90  & 84.38 & 80.87 & 82.59 \\
SE & 82.19 & 80.57 & 81.37 & 72.69 & 79.82 & 76.09 & 70.15 & 89.77 & 78.76 & 90.89 & 72.01 & 80.36 & 84.22 & 87.70  & 85.93 \\ \midrule
\textbf{AVG} &
  \textbf{86.68} &
  \textbf{86.75} &
  \textbf{86.50} &
  \textbf{76.49} &
  \textbf{75.92} &
  \textbf{75.39} &
  \textbf{80.98} &
  \textbf{81.63} &
  \textbf{80.69} &
  \textbf{79.40} &
  \textbf{79.72} &
  \textbf{79.29} &
  \textbf{84.26} &
  \textbf{84.71} &
  \textbf{84.27} \\
\bottomrule
\end{tabular}}
\caption{Recall (Rec), Precision(Pre) and F1 Performance of Dialogue Culture Classification Model based on mRoberta. The performance indicators are reported for 13 different cultural categories.}
\label{tb:ax_mroberta_cls_ret}
\end{table*}
\end{document}